\documentclass[letterpaper, 10 pt, conference]{ieeeconf}
\IEEEoverridecommandlockouts
\usepackage[utf8]{inputenc}
\usepackage[french]{babel}
\usepackage{graphicx}
\usepackage{amsmath}
\usepackage[T1]{fontenc}
\usepackage{float}
\usepackage{hyperref}
\usepackage{siunitx}
\usepackage{pdflscape}
\usepackage{multirow}
\usepackage{graphicx}  
\usepackage[font=small,labelfont=bf]{caption}
\usepackage{capt-of}  

\title{Deep Reinforcement Learning with anticipatory reward in LSTM for Collision Avoidance of Mobile Robots }

\author{Olivier Poulet$^{1,*}$, Frédéric Guinand$^{1,2}$, François Guérin$^{3}$%
\thanks{*This work was not supported by any organization}%
\thanks{$^{1}$LITIS, Le Havre Normandy University, 76600 Le Havre, France.
        \texttt{pouletolivier@yahoo.fr}}%
\thanks{$^{2}$Faculty of Mathematics and Natural Sciences,
        Cardinal Stefan Wyszyński University, 01-815 Warsaw, Poland.
        \texttt{frederic.guinand@univ-lehavre.fr}}%
\thanks{$^{3}$GREAH, Le Havre Normandy University, 76600 Le Havre, France.
        \texttt{francois.guerin@univ-lehavre.fr}}%
}

\begin{document}

\maketitle

\begin{abstract}
This article proposes a collision risk anticipation method based on short-term prediction of the agents position. A Long Short-Term Memory (LSTM) model, trained on past trajectories, is used to estimate the next position of each robot. This prediction allows us to define an anticipated collision risk by dynamically modulating the reward of a Deep Q-Learning Network (DQN) agent. The approach is tested in a constrained environment, where two robots move without communication or identifiers. Despite a limited sampling frequency (1 Hz), the results show a significant decrease of the collisions number and a stability improvement. The proposed method, which is computationally inexpensive, appears particularly attractive for implementation on embedded systems.

\end{abstract}

\section{Introduction}
In many mobile robotics scenarios, agents must collaborate or coexist in shared and constrained spaces (warehouses, corridors, urban environments…) without necessarily having explicit identifiers, communication or knowledge about the intentions or trajectories of the others \cite{raibail2022decentralized}. The case of mobile robots having identical shapes and operating without explicit identifiers or direct communication are still little studied in the scientific literature. The founding articles are the study of pattern formation \cite{suzuki1993distributed}, \cite{flocchini2014distributed}. The robots that we consider have no identifiers or distinctive shapes and work in an autonomous way. They have the same low level control algorithms and no communication between them. Each mobile robot receives at the same time (synchronization) the coordinates of the others and keeps in memory, their own previous movements. This framework makes it difficult to elect a leader \cite{dieudonne2010leader} for coordination and planning \cite{franchi2013mutual}. For the specific design of our anti-collision system, robots’ anonymity supposes to not exchange identifiers and using deep reinforcement learning \cite{chen2017decentralized} or probabilistic methods like Voronoi distances \cite{zhu2022decentralized}. Artificial intelligence is increasingly used for this purpose \cite{rafai2022review}. It improves robustness in dynamic environments and enables real-time adaptability. The techniques used rely on reinforcement learning such as Deep Q Network (DQN), policy-based methods, or actor-critic \cite{lowe2017multi}. \\

DQN requires the use of a neural network able to take into account a significant number of input measurements, such as robot's kinematic measurements, information related to the objective, and exteroceptive sensors measurements such as LiDAR \cite{chen2017decentralized}. LiDAR can provide more than one hundred measurements (impacts) to cover the entire environment. Environmental detection can also be achieved by fusing different sensors, which enhances the robustness of obstacle detection and collision avoidance systems \cite{wang2019multi} in autonomous vehicles. \\

The reward function is a crucial issue to help the agent to take the best decision. Rewards influence the effectiveness of learning algorithms \cite{liu2020reinforcement}. \\

The use of reward shaping brings an additional component that encourages or penalizes risky behaviour to avoid possible collisions \cite{zhang2020obstacle}. In the reinforcement learning context, reward shaping is a technique aimed at accelerate learning by providing additional rewards to the agent, based on a priori knowledge of the environment. A specific approach \cite{ng1999policy}  called Potential-Based Reward Shaping (PBRS) ensures that the optimal policy remains unchanged when these rewards are added. \\

The function $F(s,s')$ is defined in terms of a potential function $\Phi$ assigning a value to each state $s$:

\begin{equation}
F(s, s') = \gamma \cdot \Phi(s') - \Phi(s)
\end{equation}

where:

\begin{itemize}
\item  \( s \) and \( s' \) are the current and next states respectively,
\item \( \gamma \) is the discount factor,
\item \( \Phi \)  is the potential function.\\
\end{itemize}

Adding this reward shaping function to the original reward \( R(s, a, s') \), we obtain a new reward \( R'(s, a, s') \) defined by:

\begin{equation}
R'(s, a, s') = R(s, a, s') + F(s, s')
\end{equation}

The selection of $\Phi$ highly depends on task and environment. It relies on experience and knowledge of the environment to be properly integrated.\\

This approach ensures that the agent's global behaviour remains unchanged while accelerating the learning process by providing additional insights into the states value.\\

In this context, the use of predictive models, such as Long Short-Term Memory (LSTM) networks, and reinforcement learning strategies makes it possible to consider previous trajectories to improve travel safety and efficiency, while respecting onboard processing constraints.\\

In our previous work \cite{poulet2018self}, \cite{poulet2021experimental}, \cite{poulet2024self}, we studied the contribution of deep learning to the localization of anonymous robots compared to traditional methods. We also highlighted the used structure network and demonstrated that the anonymity of the received coordinates was not an obstacle for deep learning.\\

 Here, we propose an anticipatory method to improve the robot movements safety by integrating an estimate of the agents' next position into the reward function. This estimate is performed using a Long Short-Term Memory (LSTM) model \cite{paszek2023using}, \cite{hochreiter1997long}. The LSTM network is trained on previous trajectories, allowing it to predict next positions. These positions allow it to estimate the next mobile robots reward knowing their current trajectory.\\

The resulting prediction is then used by a Deep Q-Network (DQN) agent whose structure is similar to \cite{poulet2024self}. This DQN dynamically modulates the reward based on an estimated next collision risk. The objective is to select safe trajectories, taking into account their current shapes and probable evolution.\\

This strategy stands out by its ability to use low-frequency (1 Hz) sampled position data chosen from our constrained environment, making it suitable for embedded systems with limited resources. Experimental results show a significant reduction in the collision rate, confirming the interest of such an approach in the field of collaborative robotics.
We demonstrate the effectiveness of our method through simulations on “Webots” software \cite{michel2004cyberbotics}.\\

This article is structured as follows: Part II formally describes the proposed method and the experimental setup. Part III compares the results obtained in the initial state without anticipation and those obtained with the anticipated reward. Parts IV analyses the results. Conclusion and future works are described in part V.


\section{Proposed method and experimental setup}

\subsection{Problem formulation}

In reinforcement learning strategies, the result of an action leading to improvement and a reward (Figure~\ref{renforcement}) is usually described starting from the quality, defined in its deterministic form, as follows (equation \ref{renforce}) :

\begin{figure}
    \centering
    \includegraphics[width=0.4\textwidth]{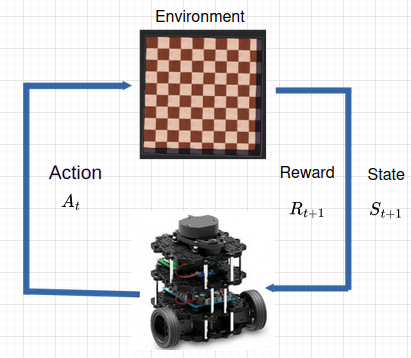}
    \caption{Interactions between a mobile robot and its environment in the reinforcement learning context.}
    \label{renforcement}
\end{figure}

\begin{equation}
Q_t(s, a) = R_t + \gamma \max_{a'} Q_{t+1}(s', a')
\label{renforce}
\end{equation}

where : 

\begin{itemize}
    
\item $Q_t(s,a)$ : Value of the quality a of state $s$ at time $t$,

\item $R_t$ : Reward received after performing the action,

\item $\gamma$ : Discount factor,

\item $s'$, $a'$ : Next state and action,

\item $\max_{a'} Q_{t+1}(s', a')$ : Best estimated quality of the next state.\\
\end{itemize}

\begin{figure}
    \centering
    \includegraphics[width=0.4\textwidth]{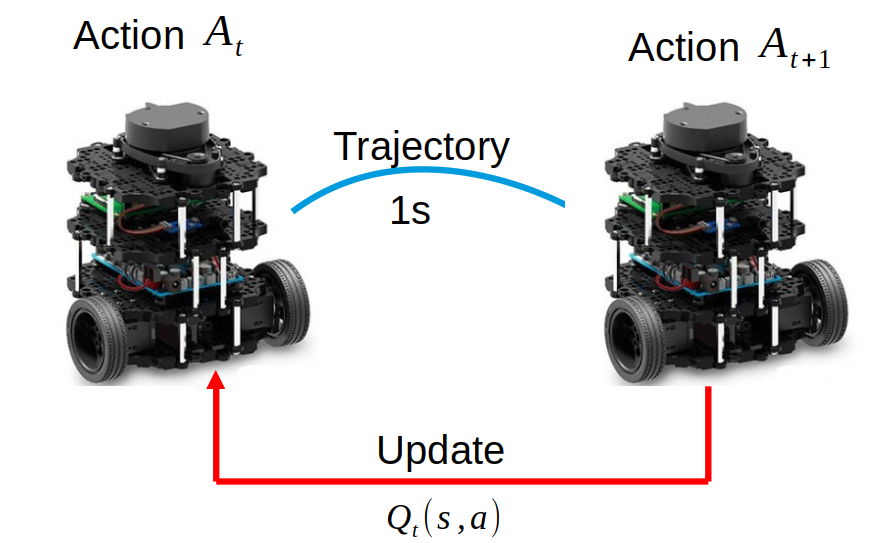}
    \caption{Evolution of the quality between two periods.}
    \label{renforcement2}
\end{figure}

\begin{figure}
    \centering
    \includegraphics[width=0.5\textwidth]{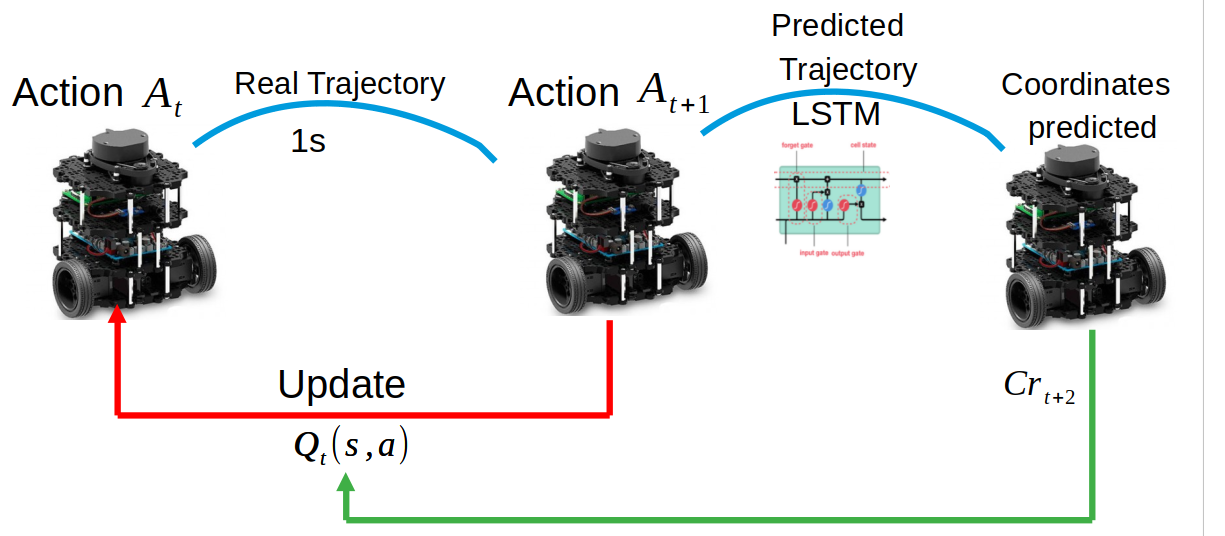}
    \caption{Penalty $Cr_{t+2}$ resulting from next position estimation by the LSTM networ}
    \label{renforcement3}
\end{figure}

When using sampled measurements (1s in our case), the reward Rt (which is derived from the consequence of the action At) is measured when the period following the action has expired, or in our case, at the exact moment when the mobile robot begins a new action. The evolution of the quality in function of time is represented on Figure~\ref{renforcement2}. We can clearly see that only once the new decision $At+1$ is made, the reward is estimated to update the quality $Q_t$. This also means that we know the upcoming action as well as the new state $s'$. Thus, we are able to estimate the next position of the mobile robots at time $t+2$ with a previously trained LSTM network. In other words, since each mobile robot knows the coordinates of the others, it can therefore use a LSTM network to estimate the next position of the others at time  $t+2$, and thus determine the inter-robot distance and calculate a next reward based on the collision risk. The "Collision Risk" ($Cr$) is defined as a reward shaping that will modulate the present quality by defining the probable next collision risk after having taken the $A_t$ action (Figure~\ref{renforcement3}).\\

We then have the new expression: 
\begin{equation}
Q_t(s, a) = R_t + \gamma \max_{a'} Q_{t+1}(s', a')+k.Cr_{t+2}
\label{renforce2}
\end{equation}

k is our discount factor.

\subsection{Experimental Setup}

To estimate the efficiency of the proposed method, the initial results must be obtained without the anticipated reward shaping. These initial results will serve as reference for the comparison with the results obtained with the LSTM anticipation.\\

\begin{figure}
    \centering
    \includegraphics[width=0.4\textwidth]{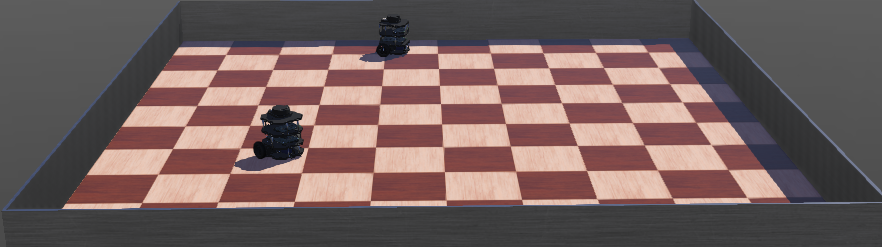}
    \caption{“Webots” simulation software}
    \label{webots}
\end{figure}

 Experimental tests require to configure an initial setup that has been implemented with Webots \cite{michel2004cyberbotics}.
The chosen mobile robots were Turtlebot3 \cite{amsters2020turtlebot} and the operating area was set to a 2.5 x 2.5 m square. This allowed the mobile robots to operate in a relatively large environment by allowing frequent collisions to occur during tests. (Figure ~\ref{webots}). \\

\subsubsection{Without Anticipation Algorithm}
\mbox{}\

The mobile robots use the same algorithm, each of them moves at (\SI{0.25}{\meter\per\second}) as long as no obstacle is detected (\SI{0.225}{\meter}). If an obstacle is detected, the mobile robot stops during 5 seconds and turns on itself until the obstacle detection disappears.\\

The control of the mobile robots is based on a deep reinforcement learning using Deep Q-Network (DQN). The DQN, embedded in the mobile robot, realizes decision making starting from a state vector including 11 variables: 
\begin{itemize}
    \item the robot's linear speed, 
    \item the minimum distances estimated by the lidar on three frontal zones distributed between -70\textdegree  and +70\textdegree,
    \item the orientation and position (x, y) of the two mobile robots being controlled, 
    \item the Euclidean inter-robot distance. \\
\end{itemize}

At each execution cycle (1s), the mobile robot collects and normalizes these data. It then transmits them to the neural network which selects an action from three possible rotation speeds[-0.15, 0 , +0.15] ~rad/s.\\

The DQN agent uses a stochastic softmax function applied to the estimated $Q(s,a)$ values. The algorithm updates (Bellman) the neural model through supervised learning on the transitions stored in a buffer.\\

The discount factor chosen is  $\gamma = 0.9$, giving more importance on next rewards. The learning rate $\alpha$ is set to $0.0075$ and remains constant during learning. These values were chosen empirically, in accordance with common practices in deep reinforcement learning. They led to good convergence results in our preliminary tests. \\

Several types of rewards were used. The best results were obtained under the following conditions:
\begin{itemize}
    \item strongly negative if the robot detects an obstacle  (\text{distance} < 0.225\ \text{m}), reward = -1,
    \item weakly negative if an obstacle is close (0.225\ \text{m} \text{distance} < 0.450\ \text{m}), reward =-0.2,
    \item strongly positive if no obstacle is detected, reward = +1.\\
\end{itemize}

The best neural network structure (determined by experimentation) consists of 3 layers of 50 neurons each. The decision process is derived from a Boltzmann distribution applied to the $Q(s,a)$ values provided by the neural network. To gradually strengthen exploitation over exploration, these values are multiplied by an factor $k_t$, initialized to 1 and incremented every $10\,000$ time steps. The action probability is then given by the following equation: \cite{cruz2018action}:

\begin{equation}
\pi(a \mid s) = \frac{ \exp\left( k_t \cdot Q(s,a) \right) }{ \sum_{a'} \exp\left( k_t \cdot Q(s,a') \right) }.
\end{equation}

This strategy leads to gradually reducing the effective temperature of the softmax distribution, making the decision process more deterministic over the training course.\\

This approach doesn’t explicitly modify the denominator of the conventional softmax function. The tests are carried out on sequences including $100\,000$ measurements, or 10 factor $k_t$ steps of change.\\

\subsubsection{With Anticipation Algorithm}
\mbox{}\

The anticipatory reward model uses a two-layer LSTM architecture, each with 50 hidden neurons and a dropout of 0.3, processing sequences of 20 time steps. Each input vector comprises 8 features (linear and angular velocity, pose robot 1, pose robot 2) and the final layer produces the following predicted positions \((x_1, y_1, x_2, y_2)\). We train with the mean squared error loss minimized by an Adam optimizer (learning rate = 0.0005, weight\_decay = 1e-5). Training was performed for 50 epochs with a batch size of 32. Inputs are normalized using their mean and standard deviation. At each interaction cycle with the environment, if the history is complete, the LSTM neural network predicts the next position at time $T+1$ and therefore calculates the anticipated distance between the two mobile robots. The objective is to estimate the dynamic aspects of the environment, i.e., the possible collisions between mobile robots. An anticipated penalty or reward is then applied based on this predicted distance, according to the following rule: 

\begin{itemize}
    \item a strong penalty if the inter-robot distance is less than \SI{0.225}{\meter}, $Cr=-0.2$ ,
    \item no reward if the inter-robot distance is between \SI{0.225}{\meter} and \SI{0.45}{\meter}, $Cr=0$,
    \item a positive reward if the distance exceeds \SI{0.45}{\meter}. $Cr=+0.2$\\
\end{itemize}

This anticipated penalty ($Cr$) is added to the immediate reward to encourage the agent to avoid configurations that potentially lead to a collision, even before it occurs. The LSTM neural network is trained on more than 130\,000 measurement points, with 80\% used for training and 20\% for testing. For both robots, the LSTM training converged very quickly, with training and validation losses approaching near-zero after only a few epochs (Figure~\ref{training}). This indicates a robust generalization to unseen data and minimal overfitting, with in a final test RMSE around 0.03. Such stability and accuracy suggest the network reliably predicts short-term robot positions.

The weighting coefficient $k$ is unitary.

\section{Experimental Results}

\subsection{Comparative Evaluation Protocol}

The LSTM and reinforcement learning approaches were implemented in Python using the PyTorch library~\cite{pytorch}. Measurements were carried out without integrating the anticipation reward mechanism (i.e., without reward shaping) in order to establish a baseline for the comparative evaluation. All validation experiments were carried out to get 100\,000 measurement points, corresponding to the same number of decisions made by each robot during the simulation.\\

The evaluation focused on several indicators estimated in segments of 10\,000 points:
\begin{itemize}
    \item the minimum number of collisions is observed per segment, allowing us to assess the agent's ability to avoid collisions over short periods,
    \item the total number of collisions between points 40\,000 and 100\,000, to highlight the long-term stabilization,
    \item the cumulative reward averaged over 100 measurements, serving as an overall performance indicator,
    \item the reward standard deviation in each segment, providing information about the variability and stability of the agent's behavior.\\
\end{itemize}

Only the best result obtained without anticipation is kept and presented in this article for comparison purposes. However, several experiments including an anticipated reward using the LSTM model with various parameters (type of penalties, intensity of predictions, distance thresholds) are explored. These different experiments makes it possible to identify the optimal conditions for using the anticipation mechanism and to precisely assess its impact on reducing collisions and improving the relevance of the decision process.\\

These measurements are systematically recorded during training to monitor performance changes and allow the comparison between both configurations (with or without anticipation). The data resulting from this phase are then used to demonstrate the impact of the reward shaping mechanism on reducing collisions.

\subsection{Results}
The results are summarized in Table~\ref{results}

For two points of the environment $P(x1,y1)$and $Q(x2,y2)$, the distances used for reward shaping are :

\begin{itemize}
    \item The Manhattan distance \cite{suwanda2020analysis} is defined by:
    \begin{equation}
    d_{\text{Manhattan}}(P, Q) = |x_1 - x_2| + |y_1 - y_2|
    \label{eq:manhattan}
    \end{equation}
    It corresponds to the distance travelled in a grid-structured space, such as traveling in an urban environment.

    \item The square of the Euclidean distance is defined by:
    \begin{equation}
    d^2_{\text{Euclidean}}(P, Q) = (x_1 - x_2)^2 + (y_1 - y_2)^2
    \label{eq:euclidean_squared}
    \end{equation}
    Because the squared distance grows quadratically, this approach naturally encourages higher inter-robot separations, thereby enhancing collision avoidance.\\
\end{itemize}

\begin{table*}[ht]
\centering
\caption{Comparison of tests with and without anticipation}
\small
\resizebox{\textwidth}{!}{%
\begin{tabular}{|c|c|cccccc|c|}
\hline
\multirow{2}{*}{\textbf{Test number}} &
\multirow{2}{*}{\textbf{Anticipation (LSTM)}} &
\multicolumn{6}{c|}{\textbf{Measurements on the optimum segment}} &
\multirow{2}{*}{\shortstack{\textbf{Total} \\ \textbf{collisions (40k--100k)}}} \\
\cline{3-8}
& & \textbf{Segment} & \textbf{Avg. Rewards R1} & \textbf{Avg. Rewards R2} & \textbf{Std Dev  R1} & \textbf{Std Dev  R2} & \textbf{Collisions R1+R2} & \\
\hline
1 & No & 70k--80k & 0.90 & 0.95 & 0.07 & 0.05 & 25 & 204 \\
2 & Yes & 90k--100k & 0.91 & 0.94 & 0.08 & 0.05 & 17  & 167 \\
3 & Yes & 60k--70k & 0.93 & 0.95 & 0.04 & 0.03 & 11 & 91 \\
4 & Yes & 70k--80k & 0.98 & 0.98 & 0.01 & 0.02 & 5  & 85 \\
\hline
\end{tabular}
}
\label{results}
\end{table*}

\begin{figure*}[htbp]
  \centering
  \includegraphics[width=0.61\textwidth]{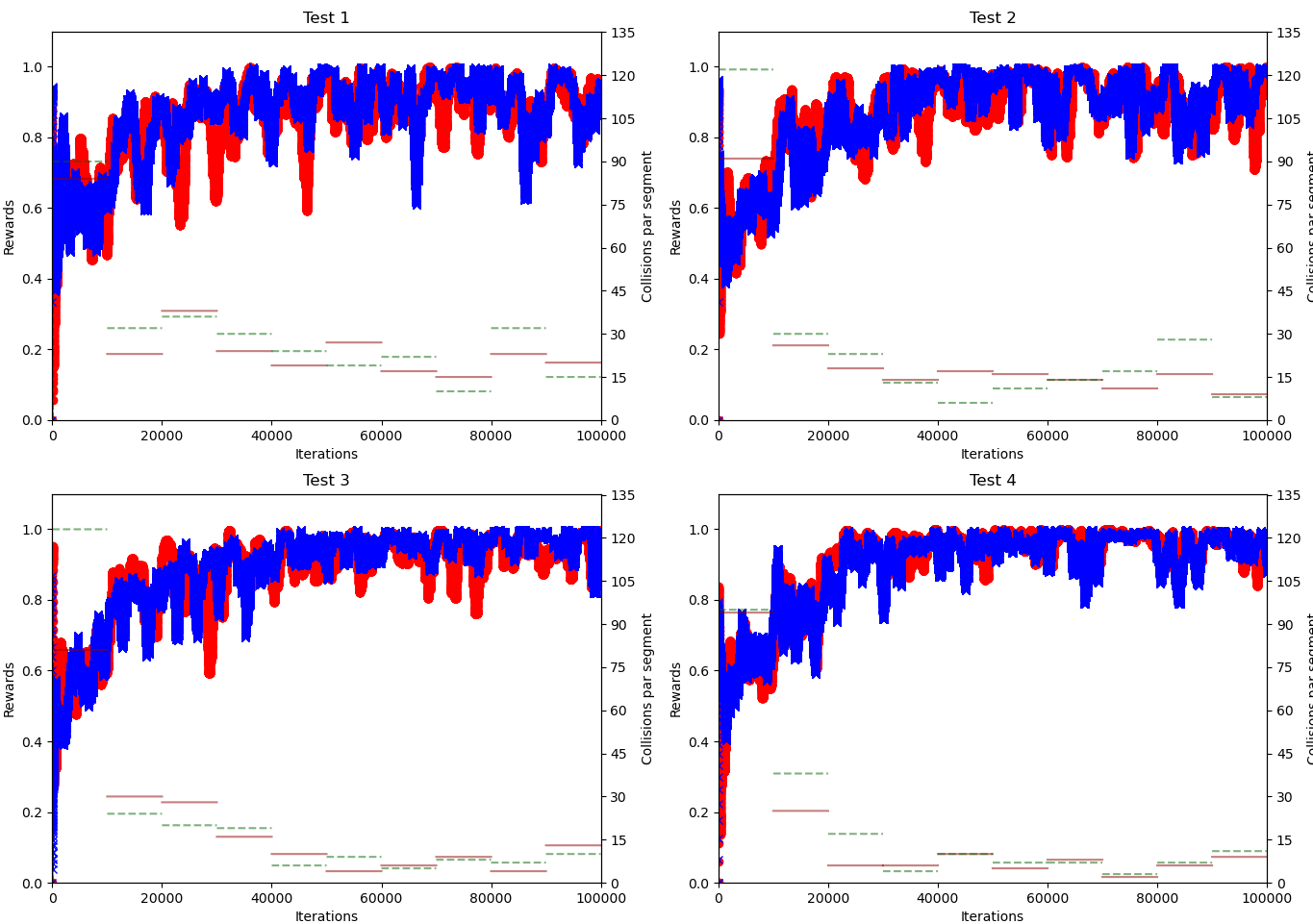}
  \caption{Comparative tests No.1, No.2, No.3, and No.4 (limited to 100000 points).
Red curve: Robot 1 – Blue curve: Robot 2.
Continuous curves correspond to the running average of the last 100 rewards.
Segments represent the number of collisions per 10,000-point segment.}
  \label{reward}
\end{figure*}

The results presented in test No.1 do not present any anticipation and serve as a reference for the following tests with anticipation. The results presented in test No.2 were obtained by considering the Manhattan distance as metric for evaluating the proximity between mobile robots. However, tests No.3 and No.4 are performed in the same configuration to highlight repeatability. They are identical to test N\textsuperscript{o}~2, except for the metric used, which is based here on the square of the Euclidean distance. The four windows in Figure~\ref{reward} correspond to tests No.1 to No.4, representing the reward evolution as a rolling average from the beginning to the end ( 100000 points) of learning, where the mobile robots can be considered mainly in operation.\\

\begin{figure*}[htbp]
  \centering
  \begin{minipage}[t]{0.34\textwidth}
    \centering
    \includegraphics[width=\linewidth]{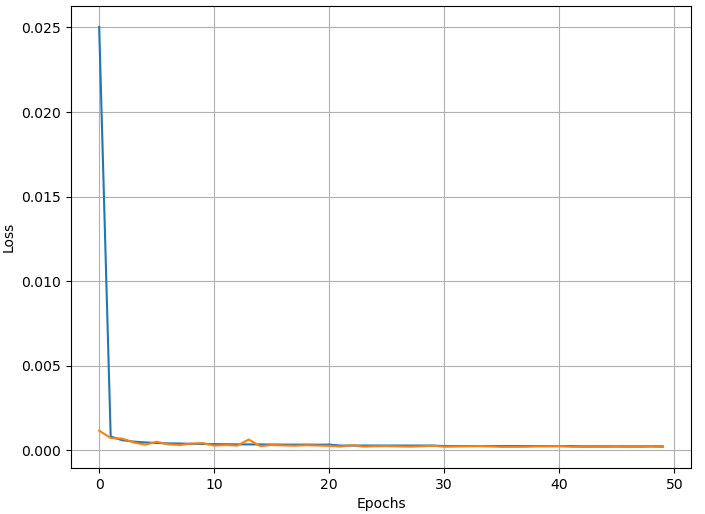}
    \captionof{figure}{Training loss (blue) and validation loss (orange) versus epochs for robot 1.}
    \label{training}
  \end{minipage}
  \hfill
  \begin{minipage}[t]{0.65\textwidth}
    \centering
    \includegraphics[width=\linewidth]{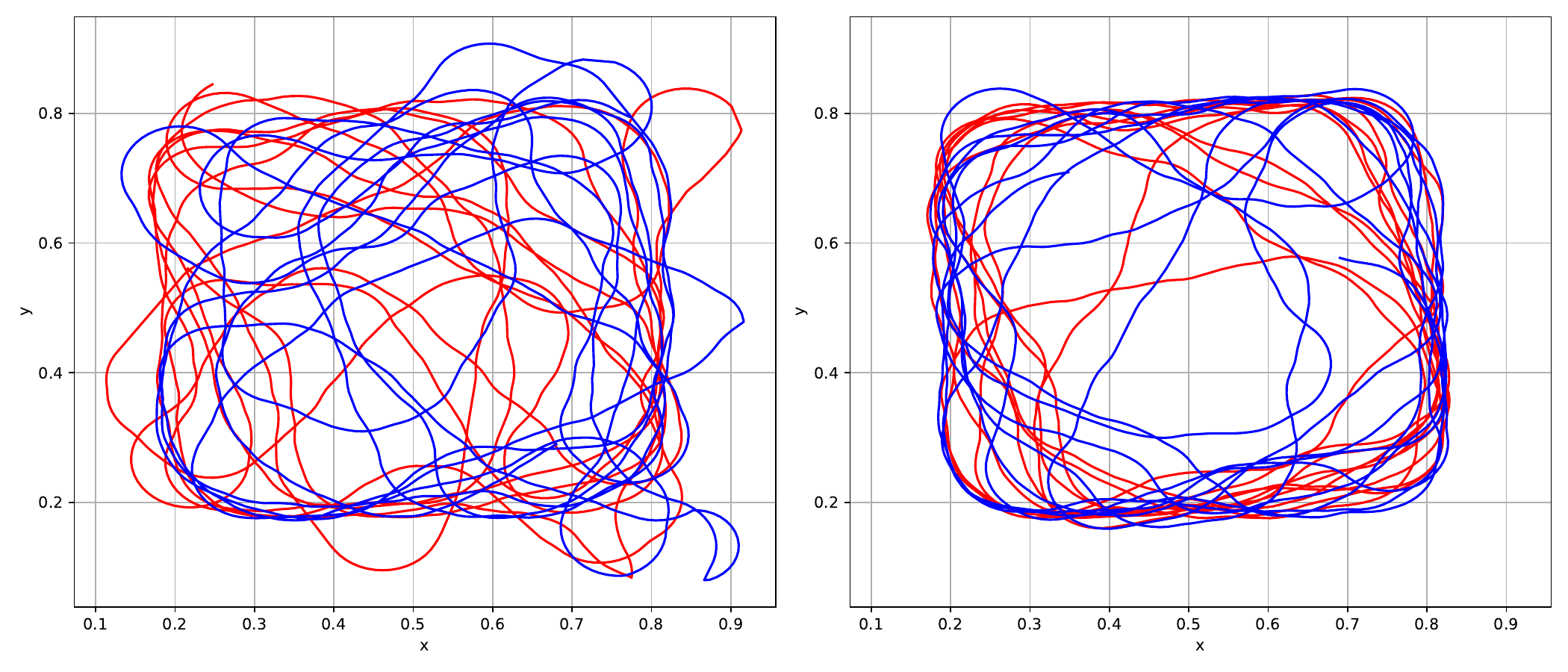}
    \captionof{figure}{View of robot 1 (red) and robot 2 (blue) trajectories for test~4.
                      2000 normalized points considered after 90000 training points.}
    \label{trajectoire}
  \end{minipage}
\end{figure*}

Figure~\ref{trajectoire} highlights the robots' movement in both tests No.2 and No.4 when the mobile robots are in operation to demonstrate the improvement of the anti-collision system.

\section{Discussion}

The experimental results obtained from the comparative evaluation of the four tested configurations highlight the contribution of anticipation in the design of the reward function. The predictive model integration (LSTM), aimed at estimating the next position of the mobile robots in short term, allows a dynamic reward adaptation resting on anticipated collision risk. This dynamic reward adaptation, similar to decision anticipation, results in significant improvements in terms of safety and stability:

\begin{itemize}
  \item Reduction of the collisions number: By incorporating a squared Euclidean distance as an anticipated penalty (Test No.4), the total number of collisions is reduced by more than 55\,\% compared to Test No.1 (without anticipation). This result validates the initial intuition that introducing a penalizing information proportional to a supposed risk in the immediate future encourages the agent to select more cautious trajectories. This can be observed in the compared trajectories of the two mobile robots.

  \item Stabilization: The addition of anticipation stabilizes the global behaviour of the mobile robots. Tests No.3 and No.4 exhibit lower standard deviations than Test No.1, reflecting more relevant decision-making. This stabilization is essential to improve implicit coordination between agents. This is also reflected in the shape of the trajectories (Figure~\ref{trajectoire}).

  \item Effectiveness of the LSTM model: The prediction model, based on an LSTM architecture using the last 20 observations, produces an efficient estimate of the next position. Its low computational cost makes it a suitable candidate for embedded implementation.

  \item Influence of the metric used: Switching from Manhattan distance (Test No.2) to squared Euclidean distance (Tests No.3 and No.4) improves the performances. This suggests that a continuous metric refines reward modulation and produces more suitable trajectories.

\item Coordinate Indexing robustness: As previously observed in the related article \cite{poulet2024self}, the absence of explicit indexing for the robots' coordinates does not cause any issues in the system's operation.

  \item Limitations: The tests were conducted in a restricted environment with two mobile robots and strong assumptions about position availability. An extension to more complex environments, including obstacles and uncertainties, is a future work opportunity.
\end{itemize}

\section{Conclusion and Future Research}

This article has presented a collision risk anticipation method based on short-term prediction of next positions using an LSTM model. This approach allows a dynamic modulation of the reward function based on the expected evolution of the environment, without prior identification of the other agents. Experimental results show a significant improvement in both safety (reduction in the number of collisions) and stability (reduction in reward variability).\\

The introduction of a continuous metric in the anticipated penalty, combined with the predictive model's ability to exploit temporal sequences, improved significantly the overall performance of the system. This work also highlights the efficiency of an anticipated reward to improve the mobile robots behaviours in shared environments without communication.\\

Several extensions are currently under investigation. The approach is being scaled to a larger number of agents and more complex environments, including intersections and crossing points with static or dynamic obstacles. This will help evaluate the robustness and scalability of the method. Integrating an uncertainty estimator into the predictions, or using variable-horizon anticipation, may lead to more cautious and adaptive behaviors. \\

Finally, experiments on a real robotic platform under partially observable conditions are planned to validate the model in cooperative robotics contexts, where behavioral anticipation is critical for safety.

\bibliographystyle{IEEEtran}

\end{document}